\documentclass[10pt,twocolumn,letterpaper]{article}

\usepackage{wacv}
\usepackage{times}
\usepackage{epsfig}
\usepackage{graphicx}
\usepackage{amsmath}
\usepackage{amssymb}

\usepackage{booktabs}
\usepackage{tabularx,ragged2e}
\usepackage{multirow, makecell}
\usepackage{color}
\usepackage[dvipsnames]{xcolor}
\usepackage{subcaption}
\usepackage{enumitem}

\def\wacvPaperID{} 

\wacvfinalcopy 

\ifwacvfinal
\def\assignedStartPage{1} 
\fi

\ifwacvfinal
\usepackage[pagebackref=true,breaklinks=true,colorlinks,bookmarks=false]{hyperref}
\else
\usepackage[pagebackref=true,breaklinks=true,colorlinks,bookmarks=false]{hyperref}
\fi

\ifwacvfinal
\else
\fi

\begin{document}

\title{HalluciNet-\textit{ing} Spatiotemporal Representations Using a 2D-CNN}

\author{Paritosh Parmar \hspace{2cm} Brendan Morris\\
University of Nevada, Las Vegas\\
{\tt\small parmap1@unlv.nevada.edu, brendan.morris@unlv.edu}
}

\maketitle

\begin{abstract}
Spatiotemporal representations learned using 3D convolutional neural networks (CNN) are currently used in state-of-the-art approaches for action related tasks. However, 3D-CNN are notorious for being memory and compute resource intensive as compared with more simple 2D-CNN architectures. We propose to hallucinate spatiotemporal representations from a 3D-CNN teacher with a 2D-CNN student.  By requiring the 2D-CNN to predict the future and intuit upcoming activity, it is encouraged to gain a deeper understanding of actions and how they evolve.  The hallucination task is treated as an auxiliary task, which can be used with any other action related task in a multitask learning setting.  Thorough experimental evaluation shows that the hallucination task indeed helps improve performance on action recognition, action quality assessment, and dynamic scene recognition tasks. From a practical standpoint, being able to hallucinate spatiotemporal representations without an actual 3D-CNN can enable deployment in resource-constrained scenarios, such as with limited computing power and/or lower bandwidth. Codebase is available here: \url{https://github.com/ParitoshParmar/HalluciNet}.
\end{abstract}
\section{Introduction}
\begin{figure*}
    \centering
    \includegraphics[width=\textwidth]{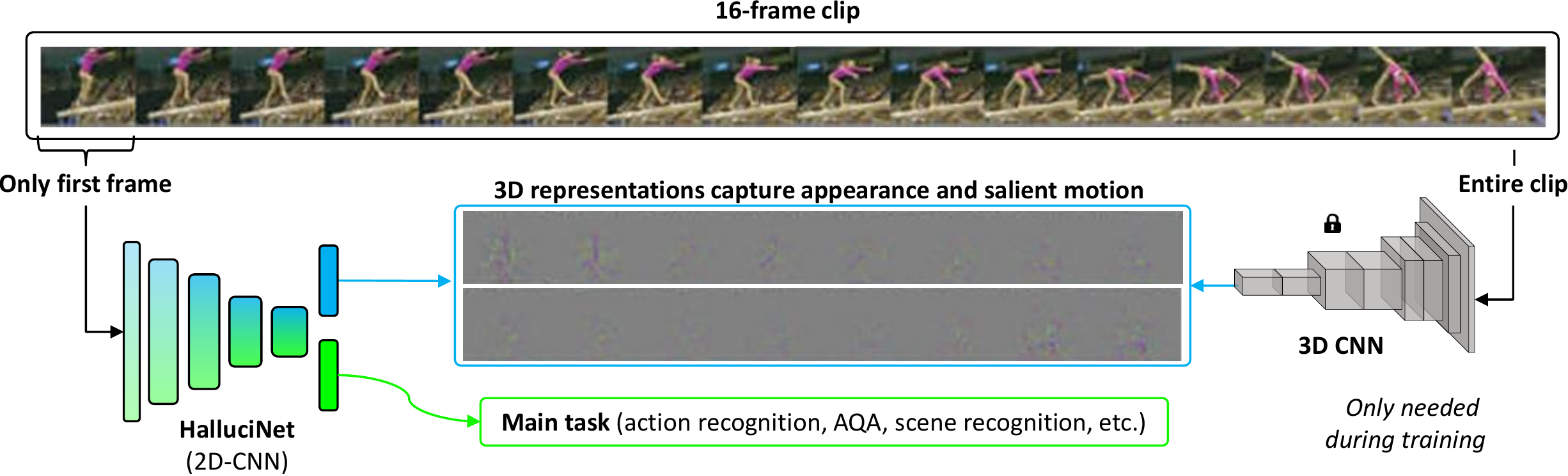}
    \caption{\textbf{Multitask leaning with HalluciNet.} HalluciNet (2D-CNN) is jointly optimized for main task, and to hallucinate spatiotemporal features (computed by an actual 3D-CNN) from a single frame.}
    \label{fig:hallucinet}
\end{figure*}

Spatiotemporal representations are densely packed with information regarding both the appearance and salient motion patterns occurring in the video clips, as illustrated in Fig. \ref{fig:hallucinet}. Due to this representational power, they are currently the best performing models on action related tasks, like action recognition \cite{c3d, kinetics, kensho, slowfast}, action quality assessment \cite{ltsoe, mtlaqa}, skills assessment \cite{doughty19}, and action detection \cite{ghanem2018activitynet}. This representation power comes at a cost of increased computational complexity \cite{mict, lightweight_wang, rgbd_zhang, characterizing_merck}, which makes 3D-CNNs unsuitable for deployment in resource-constrained scenarios.

The power of 3D-CNNs comes from their ability to attend to the salient motion patterns of a particular action class.  2D-CNNs, in contrast, are generally used for learning and extracting spatial features pertaining to a single frame/image, and thus, by design, do not take into account any motion information, therefore lacking temporal representation power. Some works \cite{walker2014patch, walker, pintea2014deja, im2flow} have addressed this by using optical flow, which will respond at all pixels that have moved/changed. This means optical flow can respond to cues both from the foreground motion of interest, as well as irrelevant activity happening in the background. This background response might not be desirable since CNNs have been shown to find \textit{short cuts} to recognize actions not from meaningful foreground, but from background cues \cite{sports1m, ar_wo_human}. These kinds of a short cuts might still be beneficial for action recognition tasks but not in a meaningful way.  That is, the 2D network is not actually learning to understand the action itself but rather contextual cues and clues.  Despite these shortcomings, 2D-CNNs are computationally lightweight, which makes them suitable for deployment on edge devices. 

In short, 2D-CNNs have the advantage of being computationally less expensive, while 3D-CNNs extract spatiotemporal features that have more representation power. In our work, we propose a way to combine the best of both worlds -- rich spatiotemporal representation, with low computational cost. Our inspiration comes from the observation that given even a single image of a scene, humans can predict how the scene might evolve. We are able to do so because of our experience and interaction in the world, which provides a general understanding of how other people are expected to behave and how objects can move or be manipulated. We propose to hallucinate spatiotemporal representations, as computed by a 3D-CNN, using a 2D-CNN, utilizing only a single still frame (see Fig. \ref{fig:hallucinet}). The idea is to force a 2D-CNN to predict the motion that will occur in the next frames, without ever having to actually see it.

\textit{Conceptually}, our hallucination task can provide richer, stronger supervisory signal that can help the 2D-CNN to gain a deeper understanding of actions and how a given scene evolves with time. Experimentally, we found our approach beneficial in following settings: 
\begin{itemize}[noitemsep]
\item actions with:
\begin{itemize}[noitemsep]
\item short-term temporal dynamics
\item long-term temporal dynamics
\end{itemize}
\item dynamic scene recognition
\item improved performance on downstream tasks when injected during pretraining
\end{itemize}
\textit{Practically}, approximating spatiotemporal features, instead of actually computing them, is useful where: 1) limited compute power (smart video camera systems, lower-end phones, or IoT devices); 2) limited/expensive bandwidth (Video Analytics Software as a Service (VA SaaS)), where our method can help reduce the transmission load by a factor of 15 (need to transmit only 1 frame out of 16). 
\section{Related Work}
Our work is related to predicting features, developing efficient/light-weight spatiotemporal network approaches, and distilling knowledge. Next, we briefly compare and contrast our approach to the most closely related works in the literature.
\paragraph{Capturing information in future frames} 
Many works have focused on capturing information in future frames \cite{yuen2010, kitani2012activity, walker2014patch, walker, pintea2014deja, finn2016unsupervised,  walker2016uncertain, koppula2015anticipating, cv_anticipating, vondrick_gen_dyna, bilen2016dynamic, vondrick_adv_xform}. Generating future frames is a difficult and complicated task, and usually require disentangling of background, foreground, low-level and high-level details and modeling them separately. Our approach of predicting features is much simpler. Moreover, our goal is not to a predict pixel-perfect future, but rather to make predictions at the semantic level.

Instead of explicitly generating future frames, works like \cite{walker2014patch, walker, pintea2014deja, im2flow} focused on learning to predict optical flow (very short-term motion information). These approaches, by design, require the use of an encoder and a decoder. Our approach does not require a decoder, which reduces the computational load. Moreover, our approach learns to hallucinate features corresponding to 16 frames, as compared to motion information in two frames. Experiments confirm the benefits of our method over optical flow prediction. 

Bilen \etal \cite{bilen2016dynamic} introduced a novel, compact representation of a video called a ``dynamic image," which can be thought of as a summary of full videos in a single image. However, computing a dynamic image requires access to all the corresponding frames, whereas HalluciNet requires processing just a single image.
\paragraph{Predicting features}
Other works \cite{actions_xforms, hoffman, cv_anticipating} propose predicting features. Our work is closest to \cite{hoffman}, where the authors proposed hallucinating depth using RGB input, whereas, we propose hallucinating spatiotemporal information. Reasoning about depth information is different from reasoning about spatiotemporal evolution.
\paragraph{Efficient Spatiotemporal Feature Computation} 
Numerous works have developed approaches to make video processing more efficient, either by reducing the required input evidence \cite{tsd, bhardwaj, mars, d3d}, or explicitly through more efficient processing \cite{sun2015human, tran2018closer, p3d, mict, xie2018rethinking, lightweight_wang, lee2018motion, tsm}. 

While these works aim to address either reducing visual evidence or developing more efficient architecture design, our solution to hallucinate (without explicitly computing) spatiotemporal representations using a 2D-CNN from a single image aims to solve both the problems, while also providing stronger supervision. In fact, our approach, which focuses on improving the backbone CNN, is complementary to some of these developments \cite{trn, tsm}.
\section{Best of Both Worlds}
Since humans are able to predict future activity and behavior through years of experience and a general understanding of ``how the world works," we would like develop a network that can understand an action in a similar manner.  To this end, we propose a teacher-student network architecture that asks a 2D-CNN to use a single frame to hallucinate (predict) 3D features pertaining to 16 frames.

Let us consider the example of a gymnast performing her routine as shown in Fig. \ref{fig:hallucinet}. In order to complete the hallucination task, the 2D-CNN should:

\begin{itemize}[noitemsep]
\item learn to identify that there's an actor in the scene and localize her;
\item spatially segment the actors and objects;
\item identify that the event is a balance beam gymnastic event and the actor is a gymnast;
\item identify that the gymnast is to attempt a cartwheel;
\item predict how she will be moving while attempting the cartwheel;
\item approximate the final position of the gymnast after 16 frames, \etc.
\end{itemize}

The challenge is understanding all the rich semantic details of the action from only a single frame.

\subsection{Hallucination Task}
The hallucination task can be seen as distilling knowledge from a better teacher network (3D-CNN), $f_t$, to a lighter student network (2D-CNN), $f_s$. The teacher, $f_t$, is pretrained and kept frozen, while the parameters of the student, $f_s$, are learned. Mid-level representations can be computed as:
\begin{align}
\phi_t &= f_t(F_0, F_1, ..., F_{T-1}) \label{eq:phi_t} \\
\phi_s &= f_s(F_0) \label{eq:phi_s}
\end{align}
where $F_T$ is the $T$-th video frame.

The hallucination loss, $\mathcal{L}_{hallu}$ encourages $f_s$ to regress $\phi_s$ to $\phi_t$ by minimizing the Euclidean distance between $\phi_s$ and $\phi_t$:
\begin{equation} \label{loss_hallu}
\mathcal{L}_{hallu} = |\sigma(\phi_s) - \sigma(\phi_t)|^2.
\end{equation}

\paragraph{Multitask learning (MTL):} Reducing computational cost with the hallucination task is not the only goal. Since the primary objective is to better understand activities and improve performance, hallucination is meant to be an auxiliary task to support the main action related task (\eg action recognition).  The main task loss (\eg, classification loss), $\mathcal{L}_{mt}$, is used in conjunction with the hallucination loss:
\begin{equation} \label{loss_overall}
\mathcal{L}_{MTL} = \mathcal{L}_{mt} + \lambda\mathcal{L}_{hallu}
\end{equation}
where, $\lambda$ is a loss balancing factor. The realization of our approach is straightforward, as presented in Fig. \ref{fig:hallucinet}.

\subsection{Stronger Supervision}
In a typical action recognition task, a network is only provided with the action class label. This may be considered a weak supervision signal since it provides a single high-level semantic interpretation of a clip filled with complex changes.  More dense labels, at lower semantic levels, are expected to provide \textbf{stronger supervisory signals}, which could improve action understanding.

In this vein, joint actor-action segmentation is an actively pursed research direction \cite{jji, gavril, yan2017, kalogeit, xu2016}.  Joint actor-action segmentation datasets \cite{a2d} provide detailed annotations, through significant annotation efforts.  In contrast, our spatiotemporal hallucination task provides detailed supervision of a similar flavor (though not exactly the same) for free.  Since 3D-CNN representations tend to focus on actors and objects, 2D-CNN can develop a better general understanding about actions through actor/object manipulation.  Additionally, the 2D representation will be less likely to take shortcuts -- ignoring the actual actor and action being performed, and instead doing recognition based on the background \cite{sports1m, ar_wo_human} -- as it cannot hallucinate spatiotemporal features, which mainly pertain to actors/foreground, from the background.

\subsection{Prediction Ambiguities}
In general, prediction of future activity with a single frame could be ambiguous (\eg, opening vs. closing a door).  However, a study has shown that humans are able to accurately predict immediate future action from a still image 85\% of the time \cite{cv_anticipating}.  So, while there may be ambiguous cases, there are many other instances where causal relationships exist and the hallucination task can be exploited.  Additionally, low-level motion cues can be used to resolve ambiguity (Sec. \ref{exp_2f}). 
\section{Experiments}
We hypothesize that incorporating the hallucination task is beneficial, by providing deeper understanding of actions. We evaluate the effect of incorporating the hallucination task in following settings:

\begin{itemize}[noitemsep]
\item (Sec. \ref{exp_ar}) Actions with short-term temporal dynamics 
\item (Sec. \ref{exp_longterm}) Actions with long-term temporal dynamics 
\item (Sec. \ref{exp_dsr}) Non-action task 
\item (Sec. \ref{exp_2f}) Hallucinating from two frames, instead of a single frame 
\item (Sec. \ref{exp_pretraining}) Effect of injecting hallucination task during pretraining 
\end{itemize} 

\noindent\textbf{Choice of networks:} In principle, any 2D- or 3D-CNNs could be used as student or teacher networks, respectively.  Noting the SOTA performance of 3D-ResNeXt-101 \cite{kensho} on action recognition, we choose to use it as our teacher network. We considered various student models. Unless otherwise mentioned, our student model is VGG11-bn, and pretrained on the ImageNet dataset \cite{imagenet}; the teacher network was trained on UCF-101 \cite{ucf101} and kept frozen. We named the 2D-CNN trained with the side-task hallucination loss as HalluciNet, and the one without hallucination loss as (vanilla) 2D-CNN, while the HalluciNet\textsubscript{direct} variant, which directly uses hallucinated features for main action recognition task. \\

\noindent\textbf{Which layer to hallucinate?} We chose to hallucinate the activations of the last bottleneck group of 3D-ResNeXt-101, which are 2048-dimensional. Representations of shallower layers will have higher dimensionality and will be less semantically mapped. \\

\noindent\textbf{Implementation details:} We used PyTorch \cite{pytorch} to implement all of the networks. Network parameters were optimized using an Adam optimizer \cite{adam} with beginning learning rate of 0.0001. $\lambda$ in Eq. \ref{loss_overall} is set to 50, unless specified otherwise. Further experiment specific details are presented with the experiment.  The codebase will be made publicly available. \\

\noindent\textbf{Performance baselines:} Our performance baseline was a 2D-CNN with same architecture, but which was trained without hallucination loss (vanilla 2D-CNN). In addition, we also compared the performance against other popular approaches from the literature, specified in each experiment. 
\begin{table}
\centering
\small
\begin{subfigure}[t]{\columnwidth}
\setlength{\tabcolsep}{6pt}
\begin{tabular}{lcc}
\toprule
\textbf{Method}                      & \textbf{UCF-static} & \textbf{HMDB-static} \\
\midrule
App stream \cite{im2flow}         & 63.60               & 35.10                \\
App stream ensemble \cite{im2flow}  & 64.00               & 35.50                \\
Motion stream \cite{im2flow}               & 24.10               & 13.90                \\
Motion stream \cite{walker}           & 14.30               & 04.96                \\
App + Motion \cite{im2flow}         & 65.50               & 37.10                \\
App + Motion \cite{walker}       & 64.50               & 35.90                \\
\midrule
Ours 2D-CNN               & 64.97               & 34.23                \\
Ours HalluciNet\textsubscript & 69.81 & \textbf{40.32} \\
Ours HalluciNet\textsubscript{direct} &	\textbf{70.53}				& 39.42 \\
\bottomrule
\end{tabular}
\caption{}
\label{tab:action_rec}
\end{subfigure}
\begin{subfigure}[t]{0.45\columnwidth}
\small
\centering
\begin{tabular}{@{}lc@{}}
\toprule
\textbf{Method}       & \textbf{Accu} \\ \midrule
TRN (R18)             & 69.57             \\
TRN (HalluciNet(R18)) & \textbf{69.94}    \\
TSM (R18)             & 71.40             \\
TSM (HalluciNet(R18)) & \textbf{73.12}    \\ \bottomrule
\end{tabular}
\caption{}
\label{tab:trn_tsm}
\end{subfigure}
\hfill
\begin{subfigure}[t]{0.45\columnwidth}
\small
\centering
\begin{tabular}{@{}lc@{}}
\toprule
\textbf{Model}             & \textbf{Accu} \\ \midrule
Resnet-50             & 76.74             \\
HalluciNet(R50) & \textbf{79.83}    \\ \bottomrule
\end{tabular}
\caption{}
\label{tab:resnet-50}
\end{subfigure}
\caption{(a) \textbf{Action recognition results and comparison}. (b) \textbf{HalluciNet helps recent developments like TRN and TSM}. (c) \textbf{Multiframe inference on better base model}. (b,c) are evaluated on UCF101.}
\end{table}
\begin{figure}
\centering
\includegraphics[width=\columnwidth]{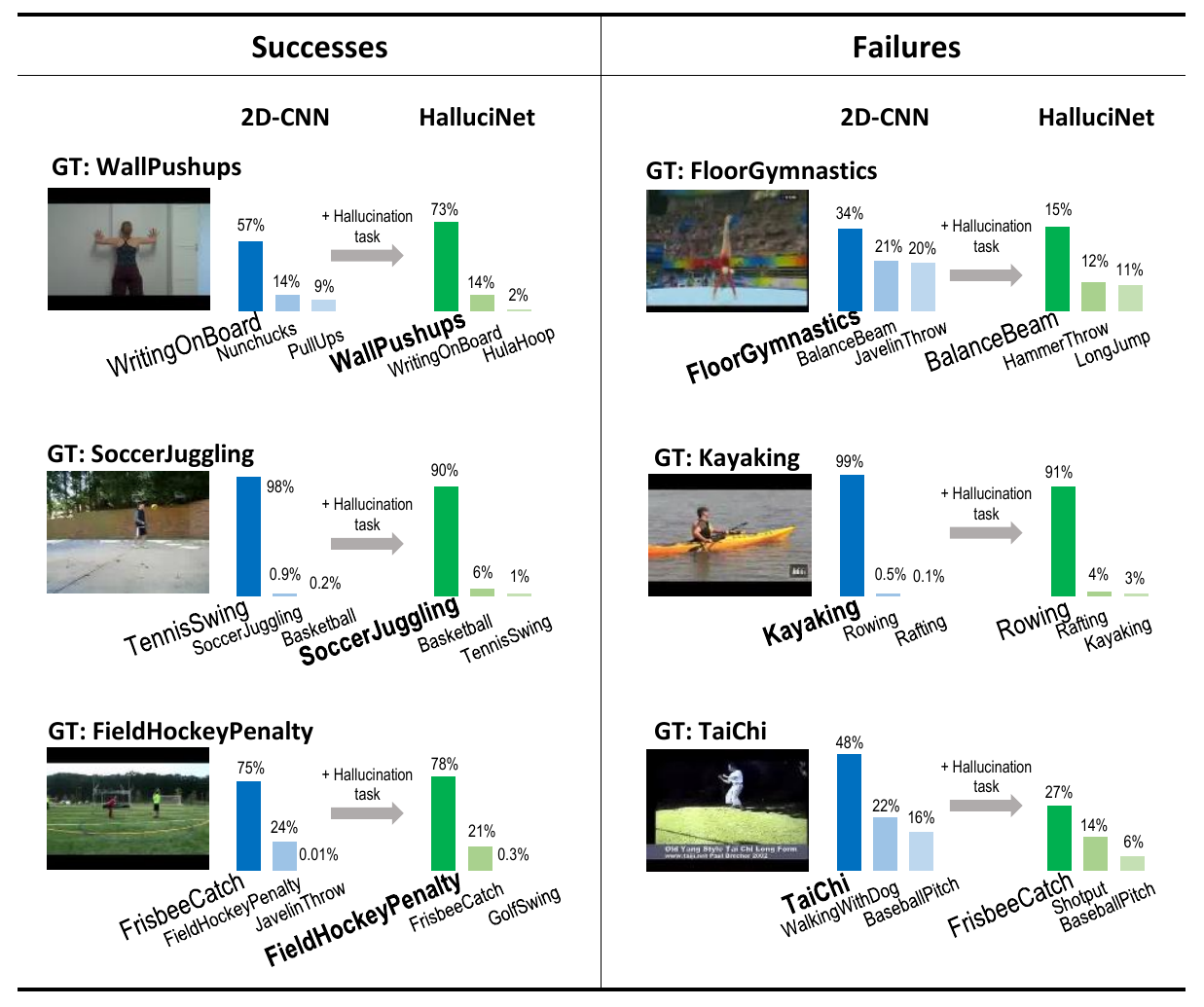}
\caption{\textbf{Qualitative results.} The hallucination task helps improve performance when the action sample is visually similar to other action classes, and a motion cue is needed to distinguish them. However, sometimes HalluciNet makes incorrect predictions when motion cue is similar to that of other actions, and dominates over the visual cue. Please zoom-in to get a better view.}
\label{qualitative_results}
\end{figure}
%
{
\setlength{\tabcolsep}{8pt}
\begin{table}
\centering
\small
\setlength\tabcolsep{3pt}
\begin{tabular}{lcccc}
\toprule
\multirow{2}{*}{\textbf{Model}} & \multicolumn{2}{c}{\textbf{Accuracy}} & \multirow{2}{*}{\textbf{\begin{tabular}[c]{@{}c@{}}Time/ \\Inf.\end{tabular}}} & \multirow{2}{*}{\textbf{\begin{tabular}[c]{@{}c@{}}\textsc{FLOP}s / \\ Param\end{tabular}}} \\ 
\cmidrule{2-3}
                                & \textbf{\textsc{\footnotesize{UCF}}}   & \textbf{\textsc{\footnotesize{HMDB}}}  &                                                                                         &                                       \\ \midrule
Ours 2D-CNN                & 64.97              & 34.23             & 3.54 ms                                                                                    & 58                                    \\ \midrule
Ours HalluciNet\textsubscript{direct}                 & 70.53              & 39.42             & 3.54 ms                                                                                    & 58                                    \\ \midrule
Ours actual 3D-CNN              & 87.95                  & 60.44                 & 58.99 ms                                                                                   & 143                                   \\ \bottomrule
\end{tabular}
\caption{\textbf{Cost vs. Accuracy Comparison.} We measure the times on Titan-X.}
\label{tab:cost_acc}
\end{table}}

\subsection{Actions with Short-Term Temporal Dynamics}
\label{exp_ar}
In the first experiment, we tested the influence of the hallucination task for general action recognition.  We compared the performance with two single frame prediction techniques: dense optical flow prediction from a static image \cite{walker}, and motion prediction from a static image \cite{im2flow}. \\

\noindent\textbf{Datasets:} UCF-101 \cite{ucf101} and HMDB-51 \cite{hmdb51} action recognition datasets were considered. In order to be consistent with the literature, we adopted their experimental protocols. Center frames from the training and testing samples were used for reporting performance, and are named as UCF- and HMDB-static, as in the literature \cite{im2flow}.

\paragraph{Metric:} We report the top-1 frame/clip-level accuracy (in \%). \\

We summarize the performance on the action recognition task in Table \ref{tab:action_rec}. We found that on both datasets, incorporating the hallucination task helped. Our HalluciNet outperformed prior approaches \cite{walker, im2flow} on both UCF-101 and HMDB-51 datasets. Moreover, our method has an advantage of being computationally lighter than \cite{im2flow}, as it does not use a flow image generator network. Qualitative results are shown in Fig. \ref{qualitative_results}. In the successes, ambiguities were resolved. The failure cases tended to confuse semantically similar classes with similar motions, such as FloorGymnastics/BalanceBeam or Kayaking/Rowing. To evaluate the quality of the hallucinated representations themselves, we directly used those representations for the main action recognition task (HalluciNet\textsubscript{direct}).  We saw that the hallucinated features had strong performance and improved on the 2D-CNN, and in fact, performed best on the UCF-static.

Next, we used hallucination task to improve the performance of recent developments TRN \cite{trn} and TSM \cite{tsm}. We use Resnet-18 (R18) as backbone for both; and implemented single, center segment, 4-frame versions of both. For TRN, we considered multiscale version. For TSM, we considered online version, which is intended for real-time processing. For both, we sample 4 frames from center 16 frames. We used $\lambda = 200$. Their vanilla versions served as our baselines. Performances on UCF101 are shown in Table \ref{tab:trn_tsm}.

We also experimented with a larger, better base model, Resnet-50. In this experiment, we trained using all the frames, and not only center frame; and during testing averaged the results over 25 frames. We used $\lambda = 350$. Results on UCF101 are shown in Table \ref{tab:resnet-50}.

Finally, in Table \ref{tab:cost_acc}, we compare our predicted spatiotemporal representation, HalluciNet\textsubscript{direct}, and the actual 3D-CNN. Hallucinet improved upon the vanilla 2D-CNN, though well below the actual 3D-CNN.  However, the performance trade-off resulted in only 6\% of the computational cost of the full 3D-CNN.

\subsection{Actions with Long-Term Temporal Dynamics}
\label{exp_longterm}
\begin{figure}
\centering
\includegraphics[width=\columnwidth]{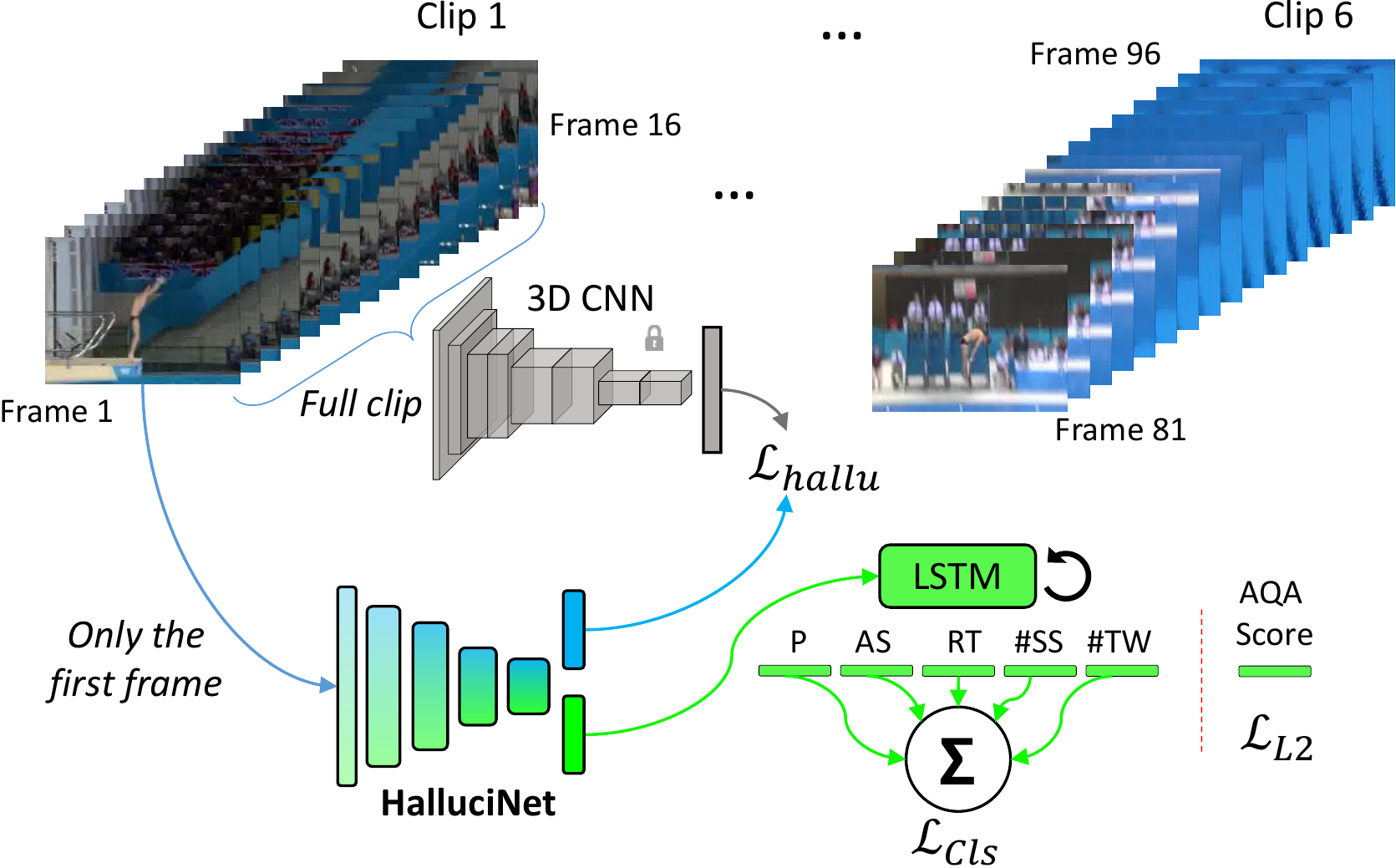}
\caption{\textbf{Detailed action recognition} and \textbf{action quality assessment models}.}
\label{fig_dar}
\end{figure}
Although we proposed hallucinating the short-term future (16 frames), frequently actions with longer temporal dynamics must be considered.  To evaluate the utility of short-term hallucination in actions with longer temporal dynamics, we considered the task of recognizing dives and assessing their quality. Short clips were aggregated over longer videos using an LSTM, as shown in Fig. \ref{fig_dar}.
\subsubsection{Dive Recognition}
\label{exp_dar}
{
\begin{table*}
\begin{subfigure}[]{0.6\textwidth}
\centering
\small
\setlength\tabcolsep{4pt}
\begin{tabularx}{\textwidth}{l c c  *5{>{\Centering}X}}
\toprule
    \textbf{Method} & \textbf{\textsc{CNN}}   & \textbf{\#Fr.} & \textbf{P} & \textbf{A} & \textbf{RT} & \textbf{SS} & \textbf{TW} \tabularnewline \midrule
C3D-AVG \cite{mtlaqa} & 3D & 96 & 96.32                 & 99.72                 & 97.45                      & 96.88                       & 93.20                  \tabularnewline
MSCADC \cite{mtlaqa} & 3D                    & 16 & 78.47                 & 97.45                 & 84.70                      & 76.20                       & 82.72                  \tabularnewline
Nibali \etal \cite{nibali} & 3D                    & 16 & 74.79                 & 98.30                 & 78.75                      & 77.34                       & 79.89                  \tabularnewline  \midrule
Ours VGG11 & \textbf{2D}               & \textbf{6} & \textbf{90.08}                 & 99.43                 & 92.07                      & 83.00                       & 86.69                  \tabularnewline
Ours HalluciNet(\footnotesize{VGG11}) & \textbf{2D} & \textbf{6} & 89.52                 & 99.43                 & \underline{\textbf{96.32}}                      & \textbf{86.12}                       & \textbf{88.10}    \tabularnewline 
Ours HalluciNet(R18) & \textbf{2D} & \textbf{6} &   \underline{\textbf{91.78}}               &     99.43             &     95.47                  &    \underline{\textbf{88.10}}                    &   \underline{\textbf{89.24}}  \tabularnewline\bottomrule             
\end{tabularx}
\caption{}
\label{tab_res_dar}
\end{subfigure}
%
%
\begin{subfigure}[]{0.4\textwidth}
\centering
\small
\setlength\tabcolsep{4pt}
\begin{tabular}{lccc}
\toprule
\textbf{Method}  &  \textbf{\textsc{CNN}} & \textbf{\#Fr.} & \textbf{Corr.} \\ \midrule
Pose+DCT \cite{pirsia}          & - & 96              & 26.82              \\
C3D-SVR \cite{ltsoe}            & 3D & 96              & 77.16              \\
C3D-LSTM \cite{ltsoe}           & 3D & 96              & 84.89              \\
C3D-AVG-STL \cite{mtlaqa}        & 3D & 96              & 89.60              \\
MSCADC-STL \cite{mtlaqa}        & 3D & 16              & 84.72              \\ \midrule
Ours VGG11        & \textbf{2D} & \textbf{6}               & 80.39                  \\
Ours HalluciNet (\footnotesize{VGG11}) & \textbf{2D} & \textbf{6}               & \textbf{82.70}                  \\ 
Ours HalluciNet (R18) & \textbf{2D} & \textbf{6}               & \underline{\textbf{83.51}}                  \\ \bottomrule
\end{tabular}
\caption{}
\label{tab_res_aqa}
\end{subfigure}
\caption{(a) \textbf{Performance comparison on dive recognition task}. \#Fr. represents the number of frames the corresponding method sees. P, AS, RT, SS, TW stand for position, arsmstand, rotation type, number of somersaults, and number of twists. (b) \textbf{Performance on AQA task}.}
\end{table*}}
\noindent\textbf{Task description:} In Olympic diving, athletes attempt many different types of dives. In a general action recognition dataset, like UCF101, all of these dives are grouped under a single action class, diving. However, these dives are different from each other in subtle ways. Each dive has the following five components: a) Position (legs straight or bent); b) starting from Armstand or not; c) Rotation direction (backwards, forwards, \etc); d) number of times the diver Somersaulted in air; and e) number of times the diver Twisted in air. Different combinations of these components produce a unique type of dive (dive number). The Dive Recognition task is to predict all five components of a dive using very few frames. \\

\noindent\textbf{Why is this task more challenging?} Unlike general action recognition datasets, \eg, UCF-101 or Kinetics \cite{kinetics}, the cues needed to identify the specific dive are distributed across the entire action sequence. In order to correctly predict the dive, the whole action sequence needs to be seen. To make the dive classification task more suitable for our HalluciNet framework, we asked the network to classify a dive correctly using only a few regularly spaced frames. In particular, we truncated a diving video to 96 frames and showed the student network every 16th frame, for a total of 6 frames.  Note that we are not asking our student network to hallucinate the entire dive sequence; rather, the student network is required to hallucinate the short-term future in order to ``fill the holes'' in the visual input datastream. \\

\noindent\textbf{Dataset:} The recently released Diving dataset MTL-AQA \cite{mtlaqa}, which has 1059 training and 353 test samples, is used for this task. The average sequence length is 3.84 seconds. \\

\noindent\textbf{Model:} We pretrained both our 3D-CNN teacher and 2D-CNN student on UCF-101.  Then the student network was trained to classify dives. Since we would be gathering evidence over six frames, we made use of an LSTM \cite{lstm} for aggregation. The LSTM was single-layered with a hidden state of 256D. The LSTM's hidden state at the last time step was passed through separate linear classification layers, one for each of the properties of a dive. The full model is illustrated in Fig. \ref{fig_dar}. The student network was trained end-to-end for 20 epochs using an Adam solver with a constant learning rate of 0.0001. We also considered HalluciNet based on Resnet-18 ($\lambda = 400$). We did not consider R50 because it is much larger compared to the dataset size.\\

The results are summarized in Table \ref{tab_res_dar}, where we also compare them with other state-of-the-art 3D-CNN based approaches \cite{nibali, mtlaqa}. Compared with the 2D baseline, Hallucinet performed better on 3 out of 5 tasks.  The \textbf{P}osition task (legs straight or bent) could be equally identifiable from a single image or clip, but the number of \textbf{TW}ists, \textbf{SS}:somersaults, or direction of rotation are more challenging without seeing motion.  In contrast, HaluciNet could predict the motion.  Our HalluciNet even outperforms 3D-CNN based approaches that use more frames (MSCADC \cite{mtlaqa} and Nibali \etal \cite{nibali}). However, C3D-AVG outperformed HalluciNet, but is computationally expensive and uses 16$\times$ more frames.

\subsubsection{Dive Quality Assessment}
\label{exp_aqa}
Action quality assessment (AQA) is another task which can highlight the utility of hallucinating spatiotemporal representations from still images using a 2D-CNN. In AQA, the task is to measure, or quantify, \textit{how well} an action was performed. A good example of AQA is that of judging Olympic events like diving, gymnastics, figure skating, \etc ~Like Dive Recognition, in order to correctly assess the quality of a dive, the entire dive sequence needs to be seen/processed. \\

\noindent\textbf{Dataset:} MTL-AQA \cite{mtlaqa}, the same as in Sec. \ref{exp_dar}.\\

\noindent\textbf{Metric:} Consistent with the literature, we report Spearman's rank correlation (in \%). 

We follow the same training procedure as in Sec. \ref{exp_dar}, except that for AQA task we used L2 loss to train, as it is a regression task. We trained for 20 epochs with Adam as a solver and annealed the learning rate by a factor of 10 every 5 epochs. We also considered HalluciNet based on R18 ($\lambda = 250$).

The AQA results are presented in Table \ref{tab_res_aqa}. Incorporating the hallucination task helped improve AQA performance. Our HalluciNet outperformed C3D-SVR and was quite close to C3D-LSTM and MSCADC although it saw 90 and 10 fewer frames, respectively.  Although it does not match C3D-AVG-STL, HalluciNet requires significantly less computation.

\subsection{Dynamic Scene Recognition}
\label{exp_dsr}
\noindent\textbf{Dataset:} Feichtenhofer \etal introduced the \textsc{YUP++} dataset \cite{yuppp} for the task of dynamic scene recognition. It has a total of 20 scene classes. The use of this dataset to evaluate the utility of inferred motion was suggested in \cite{im2flow}.  In the work by Feichtenhofer, 10\% of the samples were used for training, while the remaining 90\% of the samples were used for testing purposes. Gao \etal \cite{im2flow} formed their own split, called \textit{static-YUP++}. \\
{

\begin{table}
\centering
\small
\begin{subfigure}[t]{\columnwidth}
\centering
\small
\begin{tabular}{lr}
\toprule
\textbf{Method}         & \textbf{Accu} \\
\midrule
SFA \cite{sfa}                    & 56.90             \\
BoSE \cite{bose}                   & 77.00             \\
T-CNN \cite{tcnn}                  & 50.60             \\
\midrule
\multicolumn{2}{c}{\textit{(Ours Singleframe inference)}} \\
Ours VGG11         & 77.50             \\
Ours HalluciNet(VGG11) & \textbf{78.15}             \\
\midrule
\multicolumn{2}{c}{\textit{(Ours Multiframe inference with better base models)}} \\
Ours Resnet-50         & 83.43             \\
Ours HalluciNet(Resnet-50) & \textbf{84.44}             \\
\bottomrule
\end{tabular}
\caption{}
\label{tab:res_dyna_scene_rec_yup}
\end{subfigure}
\begin{subfigure}[t]{\columnwidth}
\centering
\small
\begin{tabular}{lc}
\toprule
\textbf{Method}              & \textbf{Accu} \\
\midrule
Appearance \cite{im2flow}                  & 74.30             \\
GT Motion \cite{im2flow}                    & 55.50             \\
Inferred Motion \cite{im2flow}              & 30.00             \\
Appearance ensemble \cite{im2flow}          & 75.20             \\
Appearance + Inferred Motion \cite{im2flow}  & 78.20             \\
Appearance + GT Motion \cite{im2flow}        & 79.60             \\
\midrule
Ours 2D-CNN              & 72.04             \\
Ours HalluciNet     & \textbf{81.53}   \\ \bottomrule         
\end{tabular}
\caption{}
\label{tab:res_dyn_scene_yup_static}
\end{subfigure}
\caption{(a) \textbf{Dynamic Scene Recognition on YUP++}. (b) \textbf{Dynamic scene recognition on static-YUP++}.}
\end{table}}

\noindent\textbf{Protocol:} For training and testing purposes, we considered the central frame of each sample. \\

The first experiment considered standard dynamic scene recognition using splits from the literature and compared them with a spatiotemporal energy based approach (BoSE), slow feature analysis (SFA) approach, and temporal CNN (T-CNN). Additionally, we also considered versions based on Resnet50 and predictions averaged over 25 frames. As shown in Table \ref{tab:res_dyna_scene_rec_yup}, HalluciNet showed minor improvement over the baseline 2D-CNN and outperformed studies in the literature. T-CNN might be the closest for comparison because it uses a stack of 10 optical flow frames; however, our HalluciNet outperformed it by a large margin. Note that we have not trained our 3D-CNN on scene recognition dataset/task, and used a 3D-CNN trained on action recognition dataset, yet still observed improvements.

The second experiment compared our approach with \cite{im2flow} in which we used their split for static-YUP++ (Table \ref{tab:res_dyn_scene_yup_static}).  In this case, our vanilla 2D-CNN did not outperform studies in literature but our HalluciNet did -- even when groundtruth motion information was used by im2flow \cite{im2flow}.  

\subsection{Using Multiple Frames to Hallucinate}
\label{exp_2f}
{\setlength{\tabcolsep}{10pt}
\begin{table*}[]
\centering
\small
\begin{subfigure}[]{0.4\textwidth}
\setlength\tabcolsep{4pt}
\begin{tabular}{llcc}
\toprule
\textbf{Method}        & \multicolumn{1}{c}{$\mathcal{L}_{hallu}$($\times$e-03)} & \textbf{Accu} & \textbf{Time/Inf.} \\
\midrule
HalluciNet(1f) & 3.3                     & 68.60         & 3.54 ms   \\
HalluciNet(2f) & 3.2 ($\downarrow3.08\%$)         & 69.55         & 5.91 ms   \\
\bottomrule
\end{tabular}
\caption{}
\label{tab:short_2f}
\end{subfigure}
\begin{subfigure}[]{0.55\textwidth}
\centering
\small
\setlength\tabcolsep{4pt}
\begin{tabular}{lccccccc}
\toprule
\multirow{2}{*}{\textbf{Method}} & \multicolumn{2}{c}{\multirow{2}{*}{$\mathcal{L}_{hallu}$($\times$e-03)}} & \multicolumn{5}{c}{\textbf{Accuracy}}                                                                             \\
\cmidrule{4-8}
& \multicolumn{2}{c}{}                                 & \textbf{P}                    & \textbf{AS}                   & \multicolumn{1}{c}{\textbf{RT}} & \multicolumn{1}{c}{\textbf{SS}} & \multicolumn{1}{c}{\textbf{TW}} \\
\midrule
HalluciNet(1f)                   & 4.2                       &                         & 89.24                & 99.72                & 94.62                  & 86.69                  & 87.54                  \\
HalluciNet(2f)                   & 3.9                       & ($\downarrow8.05\%$)                     & \textbf{92.35}       & 99.72                & \textbf{96.32}         & \textbf{89.90}         & \textbf{90.08}         \\
\bottomrule	
\end{tabular}
\caption{}
\label{tab:long_2f}
\end{subfigure}
\caption{(a) \textbf{Single-frame vs. Two-frame on \textsc{UCF-101}}. (b) \textbf{Single-frame vs. Two-frame: MTL-AQA Dive Classification}. $\mathcal{L}_{hallu}$: lower is better.}
\end{table*}
}
{\setlength{\tabcolsep}{4pt}
\begin{table}
\centering
\small
\begin{tabular}{@{}lcccccc@{}}
\toprule
\multirow{2}{*}{\textbf{Model}} &
  \multirowcell{2}{\textbf{\footnotesize{Pretraining w/}}\\\textbf{\footnotesize{Hallucination}}} &
  \multicolumn{5}{c}{\textbf{Accuracies}} \\ \cmidrule(l){3-7} 
                                 &                   & \textbf{P}     & \textbf{AS}    & \textbf{RT}    & \textbf{SS}    & \textbf{TW}    \\ \midrule
\multirow{2}{*}{2D-CNN}     & No & 90.08          & 99.43          & 92.07          & 83.00          & 86.69          \\ 
                                 & Yes  & \textbf{92.35} & \underline{\textbf{99.72}} & \textbf{94.33} & \textbf{86.97} & \textbf{88.95} \\ \midrule
\multirow{2}{*}{HalluciNet} & No & 89.52          & 99.43          & \underline{\textbf{96.32}}          & 86.12          & 88.10          \\ 
 &
  Yes &
  \underline{\textbf{92.92}} &
  \underline{\textbf{99.72}} &
  {95.18} &
  \underline{\textbf{88.39}} &
  \underline{\textbf{91.22}} \\ \bottomrule
\end{tabular}
\caption{\textbf{Utility of hallucination task in pretraining.} Best performances are underlined.}
\label{tab:hallucination_pretraining}
\end{table}}
As previously discussed, there are situations (\eg door open/close) where a single image cannot be reliably used for hallucination.  However, motions cues coming from multiple frames can be used to resolve ambiguities. 

\begin{figure}
\centering
\includegraphics[width=0.9\columnwidth]{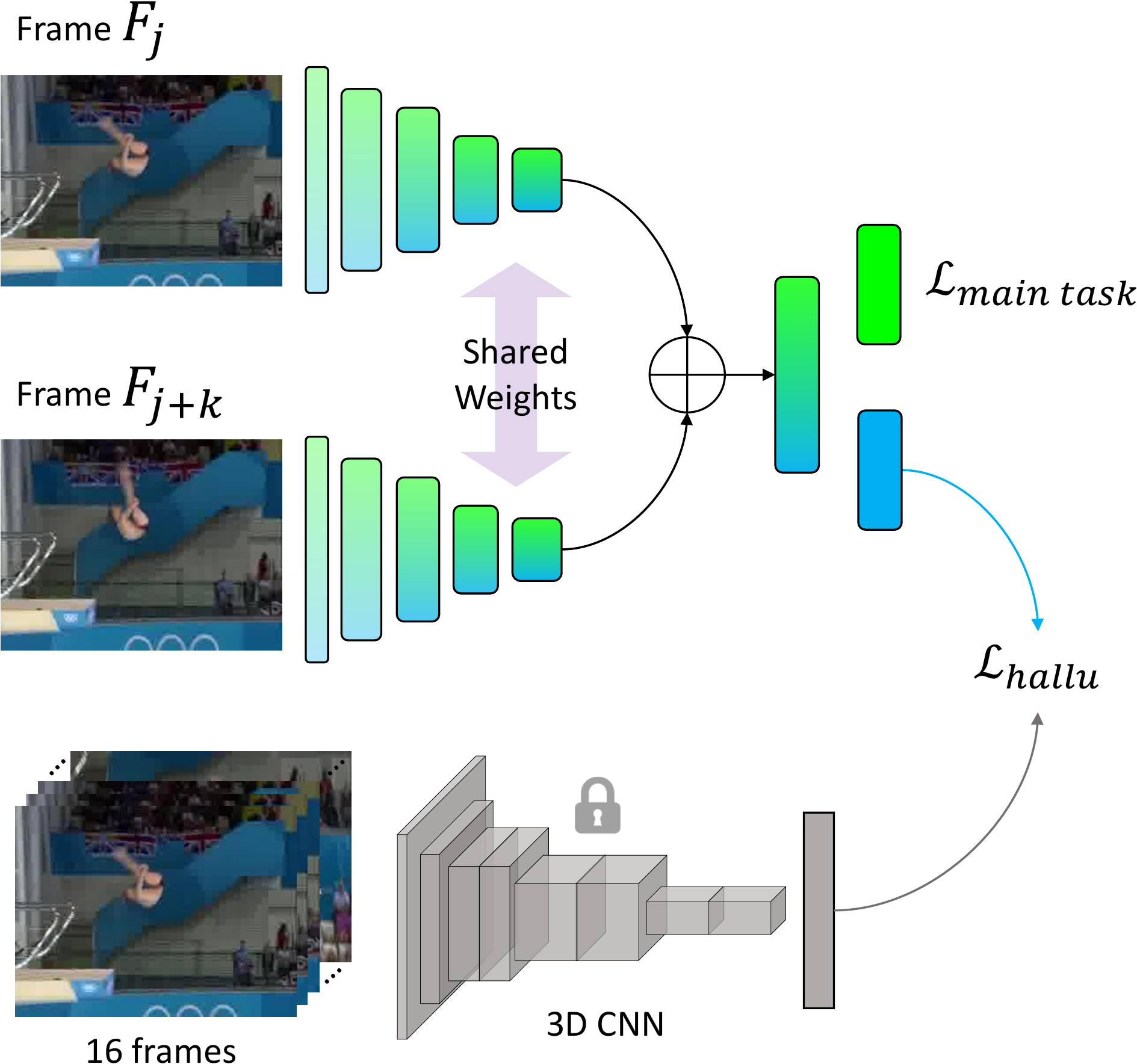} 
\caption{\textbf{Multiframe architecture}. Instead of using a single frame to hallucinate, representations of ordered frames are concatenated ($\oplus$), which is then used for hallucinating. Everything else remains same as our single frame model.}
\label{fig:multiframe_archi}
\end{figure}

We modified the single frame HalluciNet architecture to accept multiple frames, as shown in Fig. \ref{fig:multiframe_archi}. We processed frame $F_{j}$ and frame $F_{j+k}$ ($k>0$) with our student 2D-CNN. In order to tease out low-level motion cues, we did ordered concatenation of the intermediate representations, corresponding to frames $F_{j}$ and $F_{j+k}$. The concatenated student representation in the 2-frame case is: 
\begin{equation}
\phi_{s} = concat_{\phi}(\phi^{j}_{s}, \phi^{j+k}_{s})
\end{equation}
where $\phi_s^{l}$ is the student representation from frame $F_l$ as in Eq (\ref{eq:phi_s}).  This basic approach can be extended to more frames, as well as multi-scale cases. Hallucination loss remains as single frame case (Eq. \ref{loss_hallu}).

In order to see the effect of using multiple frames, we considered the following two cases:  
\begin{enumerate}[noitemsep]
\item \textbf{Single-frame baseline (HalluciNet(1f)):} We set $k=0$, which is equivalent to our standard single frame case;
\item \textbf{Two-frame baseline (HalluciNet(2f)):} We set $k=3$, to give the student network $f_{s}$ access to pixel changes in order to tease out low-level motion cues.
\end{enumerate}

We trained the networks for both the cases using the exact same procedure and parameters as in the single frame case, and observed the hallucination loss, $\mathcal{L}_{hallu}$, on the test set. We experimented with both kinds of actions -- with short-term and long-term dynamics.

Results for short-term actions are presented in Table \ref{tab:short_2f} for UCF101. We saw a reduction in hallucination loss by a little more than 3\%, which means that the hallucinated representations were closer to the true spatiotemporal representations. Similarly, there was a slight classification improvement, but with a 67\% increase in computation time.

The long-term action results are presented in Table \ref{tab:long_2f} for MTL-AQA. Like short-term actions, there was an improvement when using two frames.  The percent of reduction in $\mathcal{L}_{hallu}$ was better than the short-term case, and dive classification was improved across all components (except \textbf{AS}, which was saturated).  

\paragraph{Discussion:} Despite lower mean hallucination error in the short-term case, the reduction rate was larger for the long-term actions.  We believe this is due to the inherent difficulty of the classification task.  In UCF-101, action classes are more semantically distinguishable, which makes it easier (\eg, archery vs. applying\_makeup) to hallucinate and reason about the immediate future from a single image.  While in the MTL-AQA dive classification case, action evolution can be confusing or tricky to predict from a single image.  An example is trying to determine the direction of rotation -- it is difficult to determine if it is forward or backward with a snapshot devoid of motion. Moreover, differences between dives are more subtle.  The tasks of counting somersaults and twists need accuracy up to half a rotation.  As a result, short-term hallucination is more difficult -- it is difficult to determine if it is a full or half rotation. While the the two-frame HalluciNet can extract some low-level motion cues to resolve ambiguity, the impact is tempered in UCF-101, which has less motion dependence.  Consequently, there is comparatively more improvement in MTL-AQA, where motion (\eg, speed of rotation to distinguish between full/half rotation) is more meaningful to the classification task.  

\subsection{Utility of Hallucination Task in Pretraining}
\label{exp_pretraining}
To determine if the hallucination task positively affects pretraining, we conducted an experiment on the downstream task of dive classification on the MTL-AQA dataset. In Experiment \ref{exp_dar} the backbone network was trained on the UCF-101 action classification dataset; however, the hallucination task was not utilized during that pretraining.  Table \ref{tab:hallucination_pretraining} summarizes the results of pretraining with and without the hallucination for dive classification.  The use of hallucination during pretraining provided better initialization to both the vanilla 2D-CNN and HalluciNet, which led to improvements in almost every category besides Rotation (RT) for HalluciNet.  Additionally, HalluciNet training had the best performance for each dive class, indicating its utility both in pretraining network initialization and task-specific training.  
\section{Conclusion}
Although 3D-CNNs extract richer spatiotemporal features than the spatial features from 2D-CNNs, this comes at a considerably higher computational cost. We proposed a simple solution to approximate (hallucinate) spatiotemporal  representations (computed by a 3D-CNN) using a computationally lightweight 2D-CNN with a single frame. Hallucinating spatiotemporal representations, instead of actually computing them, dramatically lowers the computational cost (only 6\% of 3D-CNN time in our experiments), which makes deployment on edge devices feasible. In addition, by using only a single frame, rather than 16, the communication bandwidth requirements are lowered. Besides these practical benefits, we found that hallucination task when used in a multitask learning setting provides a strong supervisory signal, which helps in: 1) actions with short- and long-term dynamics; 2) dynamic scene recognition (non-action task); and 3) improving pretraining for downstream tasks. We showed that hallucination task across various base CNNs. Our hallucination task is a plug-and-play module, and we suggest future works to leverage hallucination task for action as well as non-action tasks.

{\small
\bibliographystyle{ieee_fullname}
\bibliography{egbib}

\begin{thebibliography}{10}\itemsep=-1pt

\bibitem{bhardwaj}
Shweta Bhardwaj, Mukundhan Srinivasan, and Mitesh~M Khapra.
\newblock Efficient video classification using fewer frames.
\newblock In {\em Proceedings of the IEEE Conference on Computer Vision and
  Pattern Recognition}, pages 354--363, 2019.

\bibitem{bilen2016dynamic}
Hakan Bilen, Basura Fernando, Efstratios Gavves, Andrea Vedaldi, and Stephen
  Gould.
\newblock Dynamic image networks for action recognition.
\newblock In {\em Proceedings of the IEEE Conference on Computer Vision and
  Pattern Recognition}, pages 3034--3042, 2016.

\bibitem{mars}
Nieves Crasto, Philippe Weinzaepfel, Karteek Alahari, and Cordelia Schmid.
\newblock {MARS: Motion-Augmented RGB Stream for Action Recognition}.
\newblock In {\em CVPR}, 2019.

\bibitem{imagenet}
Jia Deng, Wei Dong, Richard Socher, Li-Jia Li, Kai Li, and Li Fei-Fei.
\newblock Imagenet: A large-scale hierarchical image database.
\newblock In {\em 2009 IEEE conference on computer vision and pattern
  recognition}, pages 248--255. Ieee, 2009.

\bibitem{doughty19}
Hazel Doughty, Walterio Mayol-Cuevas, and Dima Damen.
\newblock The pros and cons: Rank-aware temporal attention for skill
  determination in long videos.
\newblock In {\em Proceedings of the IEEE Conference on Computer Vision and
  Pattern Recognition}, pages 7862--7871, 2019.

\bibitem{slowfast}
Christoph Feichtenhofer, Haoqi Fan, Jitendra Malik, and Kaiming He.
\newblock Slowfast networks for video recognition.
\newblock In {\em Proceedings of the IEEE International Conference on Computer
  Vision}, pages 6202--6211, 2019.

\bibitem{bose}
Christoph Feichtenhofer, Axel Pinz, and Richard~P Wildes.
\newblock Bags of spacetime energies for dynamic scene recognition.
\newblock In {\em Proceedings of the IEEE Conference on Computer Vision and
  Pattern Recognition}, pages 2681--2688, 2014.

\bibitem{yuppp}
Christoph Feichtenhofer, Axel Pinz, and Richard~P Wildes.
\newblock Temporal residual networks for dynamic scene recognition.
\newblock In {\em Proceedings of the IEEE Conference on Computer Vision and
  Pattern Recognition}, pages 4728--4737, 2017.

\bibitem{finn2016unsupervised}
Chelsea Finn, Ian Goodfellow, and Sergey Levine.
\newblock Unsupervised learning for physical interaction through video
  prediction.
\newblock In {\em Advances in neural information processing systems}, pages
  64--72, 2016.

\bibitem{im2flow}
Ruohan Gao, Bo Xiong, and Kristen Grauman.
\newblock Im2flow: Motion hallucination from static images for action
  recognition.
\newblock In {\em Proceedings of the IEEE Conference on Computer Vision and
  Pattern Recognition}, pages 5937--5947, 2018.

\bibitem{gavril}
Kirill Gavrilyuk, Amir Ghodrati, Zhenyang Li, and Cees~GM Snoek.
\newblock Actor and action video segmentation from a sentence.
\newblock In {\em Proceedings of the IEEE Conference on Computer Vision and
  Pattern Recognition}, pages 5958--5966, 2018.

\bibitem{ghanem2018activitynet}
Bernard Ghanem, Juan~Carlos Niebles, Cees Snoek, Fabian~Caba Heilbron, Humam
  Alwassel, Victor Escorcia, Ranjay Khrisna, Shyamal Buch, and Cuong~Duc Dao.
\newblock The activitynet large-scale activity recognition challenge 2018
  summary.
\newblock {\em arXiv preprint arXiv:1808.03766}, 2018.

\bibitem{characterizing_merck}
Ramyad Hadidi, Jiashen Cao, Yilun Xie, Bahar Asgari, Tushar Krishna, and
  Hyesoon Kim.
\newblock Characterizing the deployment of deep neural networks on commercial
  edge devices.
\newblock In {\em IEEE International Symposium on Workload Characterization},
  2019.

\bibitem{kensho}
Kensho Hara, Hirokatsu Kataoka, and Yutaka Satoh.
\newblock Can spatiotemporal 3d cnns retrace the history of 2d cnns and
  imagenet?
\newblock In {\em Proceedings of the IEEE Conference on Computer Vision and
  Pattern Recognition (CVPR)}, pages 6546--6555, 2018.

\bibitem{ar_wo_human}
Yun He, Soma Shirakabe, Yutaka Satoh, and Hirokatsu Kataoka.
\newblock Human action recognition without human.
\newblock In {\em European Conference on Computer Vision}, pages 11--17.
  Springer, 2016.

\bibitem{lstm}
Sepp Hochreiter and J{\"u}rgen Schmidhuber.
\newblock Long short-term memory.
\newblock {\em Neural computation}, 9(8):1735--1780, 1997.

\bibitem{hoffman}
Judy Hoffman, Saurabh Gupta, and Trevor Darrell.
\newblock Learning with side information through modality hallucination.
\newblock In {\em Proceedings of the IEEE Conference on Computer Vision and
  Pattern Recognition}, pages 826--834, 2016.

\bibitem{jji}
Jingwei Ji, Shyamal Buch, Alvaro Soto, and Juan Carlos~Niebles.
\newblock End-to-end joint semantic segmentation of actors and actions in
  video.
\newblock In {\em Proceedings of the European Conference on Computer Vision
  (ECCV)}, pages 702--717, 2018.

\bibitem{kalogeit}
Vicky Kalogeiton, Philippe Weinzaepfel, Vittorio Ferrari, and Cordelia Schmid.
\newblock Joint learning of object and action detectors.
\newblock In {\em Proceedings of the IEEE International Conference on Computer
  Vision}, pages 4163--4172, 2017.

\bibitem{sports1m}
Andrej Karpathy, George Toderici, Sanketh Shetty, Thomas Leung, Rahul
  Sukthankar, and Li Fei-Fei.
\newblock Large-scale video classification with convolutional neural networks.
\newblock In {\em Proceedings of the IEEE conference on Computer Vision and
  Pattern Recognition}, pages 1725--1732, 2014.

\bibitem{kinetics}
Will Kay, Joao Carreira, Karen Simonyan, Brian Zhang, Chloe Hillier, Sudheendra
  Vijayanarasimhan, Fabio Viola, Tim Green, Trevor Back, Paul Natsev, et~al.
\newblock The kinetics human action video dataset.
\newblock {\em arXiv preprint arXiv:1705.06950}, 2017.

\bibitem{adam}
Diederik~P Kingma and Jimmy Ba.
\newblock Adam: A method for stochastic optimization.
\newblock {\em arXiv preprint arXiv:1412.6980}, 2014.

\bibitem{kitani2012activity}
Kris~M Kitani, Brian~D Ziebart, James~Andrew Bagnell, and Martial Hebert.
\newblock Activity forecasting.
\newblock In {\em European Conference on Computer Vision}, pages 201--214.
  Springer, 2012.

\bibitem{koppula2015anticipating}
Hema~S Koppula and Ashutosh Saxena.
\newblock Anticipating human activities using object affordances for reactive
  robotic response.
\newblock {\em IEEE transactions on pattern analysis and machine intelligence},
  38(1):14--29, 2015.

\bibitem{hmdb51}
H. Kuehne, H. Jhuang, E. Garrote, T. Poggio, and T. Serre.
\newblock {HMDB}: a large video database for human motion recognition.
\newblock In {\em Proceedings of the International Conference on Computer
  Vision (ICCV)}, 2011.

\bibitem{lee2018motion}
Myunggi Lee, Seungeui Lee, Sungjoon Son, Gyutae Park, and Nojun Kwak.
\newblock Motion feature network: Fixed motion filter for action recognition.
\newblock In {\em Proceedings of the European Conference on Computer Vision
  (ECCV)}, pages 387--403, 2018.

\bibitem{tsm}
Ji Lin, Chuang Gan, and Song Han.
\newblock Tsm: Temporal shift module for efficient video understanding.
\newblock In {\em Proceedings of the IEEE International Conference on Computer
  Vision}, pages 7083--7093, 2019.

\bibitem{nibali}
Aiden Nibali, Zhen He, Stuart Morgan, and Daniel Greenwood.
\newblock Extraction and classification of diving clips from continuous video
  footage.
\newblock In {\em 2017 IEEE Conference on Computer Vision and Pattern
  Recognition Workshops (CVPRW)}, pages 94--104. IEEE, 2017.

\bibitem{mtlaqa}
Paritosh Parmar and Brendan~Tran Morris.
\newblock What and how well you performed? a multitask learning approach to
  action quality assessment.
\newblock In {\em Proceedings of the IEEE Conference on Computer Vision and
  Pattern Recognition}, pages 304--313, 2019.

\bibitem{ltsoe}
Paritosh Parmar and Brendan Tran~Morris.
\newblock Learning to score olympic events.
\newblock In {\em proceedings of the IEEE Conference on Computer Vision and
  Pattern Recognition Workshops}, pages 20--28, 2017.

\bibitem{pytorch}
Adam Paszke, Sam Gross, Soumith Chintala, Gregory Chanan, Edward Yang, Zachary
  DeVito, Zeming Lin, Alban Desmaison, Luca Antiga, and Adam Lerer.
\newblock Automatic differentiation in pytorch.
\newblock 2017.

\bibitem{pintea2014deja}
Silvia~L Pintea, Jan~C van Gemert, and Arnold~WM Smeulders.
\newblock D{\'e}ja vu.
\newblock In {\em European Conference on Computer Vision}, pages 172--187.
  Springer, 2014.

\bibitem{pirsia}
Hamed Pirsiavash, Carl Vondrick, and Antonio Torralba.
\newblock Assessing the quality of actions.
\newblock In {\em European Conference on Computer Vision}, pages 556--571.
  Springer, 2014.

\bibitem{p3d}
Zhaofan Qiu, Ting Yao, and Tao Mei.
\newblock Learning spatio-temporal representation with pseudo-3d residual
  networks.
\newblock In {\em proceedings of the IEEE International Conference on Computer
  Vision}, pages 5533--5541, 2017.

\bibitem{tcnn}
Karen Simonyan and Andrew Zisserman.
\newblock Two-stream convolutional networks for action recognition in videos.
\newblock In {\em Advances in neural information processing systems}, pages
  568--576, 2014.

\bibitem{ucf101}
Khurram Soomro, Amir~Roshan Zamir, and Mubarak Shah.
\newblock Ucf101: A dataset of 101 human actions classes from videos in the
  wild.
\newblock {\em arXiv preprint arXiv:1212.0402}, 2012.

\bibitem{d3d}
Jonathan~C Stroud, David~A Ross, Chen Sun, Jia Deng, and Rahul Sukthankar.
\newblock D3d: Distilled 3d networks for video action recognition.
\newblock {\em arXiv preprint arXiv:1812.08249}, 2018.

\bibitem{sun2015human}
Lin Sun, Kui Jia, Dit-Yan Yeung, and Bertram~E Shi.
\newblock Human action recognition using factorized spatio-temporal
  convolutional networks.
\newblock In {\em Proceedings of the IEEE international conference on computer
  vision}, pages 4597--4605, 2015.

\bibitem{sfa}
Christian Theriault, Nicolas Thome, and Matthieu Cord.
\newblock Dynamic scene classification: Learning motion descriptors with slow
  features analysis.
\newblock In {\em Proceedings of the IEEE Conference on Computer Vision and
  Pattern Recognition}, pages 2603--2610, 2013.

\bibitem{c3d}
Du Tran, Lubomir Bourdev, Rob Fergus, Lorenzo Torresani, and Manohar Paluri.
\newblock Learning spatiotemporal features with 3d convolutional networks.
\newblock In {\em Proceedings of the IEEE international conference on computer
  vision}, pages 4489--4497, 2015.

\bibitem{tran2018closer}
Du Tran, Heng Wang, Lorenzo Torresani, Jamie Ray, Yann LeCun, and Manohar
  Paluri.
\newblock A closer look at spatiotemporal convolutions for action recognition.
\newblock In {\em Proceedings of the IEEE conference on Computer Vision and
  Pattern Recognition}, pages 6450--6459, 2018.

\bibitem{cv_anticipating}
Carl Vondrick, Hamed Pirsiavash, and Antonio Torralba.
\newblock Anticipating visual representations from unlabeled video.
\newblock In {\em Proceedings of the IEEE Conference on Computer Vision and
  Pattern Recognition}, pages 98--106, 2016.

\bibitem{vondrick_gen_dyna}
Carl Vondrick, Hamed Pirsiavash, and Antonio Torralba.
\newblock Generating videos with scene dynamics.
\newblock In {\em Advances In Neural Information Processing Systems}, pages
  613--621, 2016.

\bibitem{vondrick_adv_xform}
Carl Vondrick and Antonio Torralba.
\newblock Generating the future with adversarial transformers.
\newblock In {\em Proceedings of the IEEE Conference on Computer Vision and
  Pattern Recognition}, pages 1020--1028, 2017.

\bibitem{walker2016uncertain}
Jacob Walker, Carl Doersch, Abhinav Gupta, and Martial Hebert.
\newblock An uncertain future: Forecasting from static images using variational
  autoencoders.
\newblock In {\em European Conference on Computer Vision}, pages 835--851.
  Springer, 2016.

\bibitem{walker2014patch}
Jacob Walker, Abhinav Gupta, and Martial Hebert.
\newblock Patch to the future: Unsupervised visual prediction.
\newblock In {\em Proceedings of the IEEE conference on Computer Vision and
  Pattern Recognition}, pages 3302--3309, 2014.

\bibitem{walker}
Jacob Walker, Abhinav Gupta, and Martial Hebert.
\newblock Dense optical flow prediction from a static image.
\newblock In {\em Proceedings of the IEEE International Conference on Computer
  Vision}, pages 2443--2451, 2015.

\bibitem{lightweight_wang}
Haonan Wang, Jun Lin, and Zhongfeng Wang.
\newblock Design light-weight 3d convolutional networks for video recognition
  temporal residual, fully separable block, and fast algorithm.
\newblock {\em arXiv preprint arXiv:1905.13388}, 2019.

\bibitem{actions_xforms}
Xiaolong Wang, Ali Farhadi, and Abhinav Gupta.
\newblock Actions\~{} transformations.
\newblock In {\em Proceedings of the IEEE conference on Computer Vision and
  Pattern Recognition}, pages 2658--2667, 2016.

\bibitem{xie2018rethinking}
Saining Xie, Chen Sun, Jonathan Huang, Zhuowen Tu, and Kevin Murphy.
\newblock Rethinking spatiotemporal feature learning: Speed-accuracy trade-offs
  in video classification.
\newblock In {\em Proceedings of the European Conference on Computer Vision
  (ECCV)}, pages 305--321, 2018.

\bibitem{xu2016}
Chenliang Xu and Jason~J Corso.
\newblock Actor-action semantic segmentation with grouping process models.
\newblock In {\em Proceedings of the IEEE Conference on Computer Vision and
  Pattern Recognition}, pages 3083--3092, 2016.

\bibitem{a2d}
Chenliang Xu, Shao-Hang Hsieh, Caiming Xiong, and Jason~J Corso.
\newblock Can humans fly? action understanding with multiple classes of actors.
\newblock In {\em Proceedings of the IEEE Conference on Computer Vision and
  Pattern Recognition}, pages 2264--2273, 2015.

\bibitem{yan2017}
Yan Yan, Chenliang Xu, Dawen Cai, and Jason~J Corso.
\newblock Weakly supervised actor-action segmentation via robust multi-task
  ranking.
\newblock In {\em Proceedings of the IEEE Conference on Computer Vision and
  Pattern Recognition}, pages 1298--1307, 2017.

\bibitem{yuen2010}
Jenny Yuen and Antonio Torralba.
\newblock A data-driven approach for event prediction.
\newblock In {\em European Conference on Computer Vision}, pages 707--720.
  Springer, 2010.

\bibitem{rgbd_zhang}
Haokui Zhang, Ying Li, Peng Wang, Yu Liu, and Chunhua Shen.
\newblock Rgb-d based action recognition with light-weight 3d convolutional
  networks.
\newblock {\em arXiv preprint arXiv:1811.09908}, 2018.

\bibitem{tsd}
Zhaoyang Zhang, Zhanghui Kuang, Ping Luo, Litong Feng, and Wei Zhang.
\newblock Temporal sequence distillation: Towards few-frame action recognition
  in videos.
\newblock In {\em 2018 ACM Multimedia Conference on Multimedia Conference},
  pages 257--264. ACM, 2018.

\bibitem{trn}
Bolei Zhou, Alex Andonian, Aude Oliva, and Antonio Torralba.
\newblock Temporal relational reasoning in videos.
\newblock In {\em Proceedings of the European Conference on Computer Vision
  (ECCV)}, pages 803--818, 2018.

\bibitem{mict}
Yizhou Zhou, Xiaoyan Sun, Zheng-Jun Zha, and Wenjun Zeng.
\newblock Mict: Mixed 3d/2d convolutional tube for human action recognition.
\newblock In {\em Proceedings of the IEEE Conference on Computer Vision and
  Pattern Recognition}, pages 449--458, 2018.

\end{thebibliography}
}

\end{document}